\documentclass[10pt,conference,a4paper,twocolumn] {IEEEtran}

\usepackage{graphicx}
\usepackage{epic}
\usepackage{figsize}
\usepackage{float}
\usepackage{hyperref}
\usepackage[section]{placeins}
\usepackage{cite}
\usepackage{amsmath}
\usepackage{amssymb}
\usepackage{amsfonts}
\usepackage{algorithm}
\usepackage{algpseudocode}
\newcommand{\email}[1]{{\fontfamily{cmtt}\selectfont#1\normalfont}}

\IEEEoverridecommandlockouts

\title{Semi-Supervised Anomaly Detection for the Determination of Vehicle Hijacking Tweets}

\author{\IEEEauthorblockN{Taahir Aiyoob Patel, Clement N. Nyirenda}
\IEEEauthorblockA{\emph{Department of Computer Science,}\\
\emph{Faculty of Natural Sciences,}\\
\emph{The University of the Western Cape}\\
\emph{Bellville, Cape Town 7535} \\
\emph{South Africa} \\
$^1$\email{3759655@myuwc.ac.za}\\
$^2$\email{cnyirenda@uwc.ac.za}
}}

\begin{document}

\maketitle

\begin{abstract}
In South Africa, there is an ever-growing issue of vehicle hijackings. This leads to travellers constantly being in fear of becoming a victim to such an incident. This work presents a new semi-supervised approach to using tweets to identify hijacking incidents by using unsupervised anomaly detection algorithms. Tweets consisting of the keyword "hijacking" are obtained, stored, and processed using the term frequency-inverse document frequency (TF-IDF) and further analyzed by using two anomaly detection algorithms: 1) K-Nearest Neighbour (KNN); 2) Cluster Based Outlier Factor (CBLOF). The comparative evaluation showed that the KNN method produced an accuracy of 89\%, whereas the CBLOF produced an accuracy of 90\%. The CBLOF method was also able to obtain a F1-Score of 0.8, whereas the KNN produced a 0.78. Therefore, there is a slight difference between the two approaches, in favour of CBLOF, which has been selected as a preferred unsupervised method for the determination of relevant hijacking tweets. In future, a comparison will be done between supervised learning methods and the unsupervised methods presented in this work on larger dataset. Optimisation mechanisms will also be employed in order to increase the overall performance.
\end{abstract}

\vspace{0.2cm}
\begin{IEEEkeywords}
Twitter, Vehicle Hijacking, Anomaly Detection, Unsupervised Learning, Semi-Supervised
\end{IEEEkeywords}

\section{Introduction}
It was reported that in 2022, there was a total of 23,025 hijackings reported, which averages to near 63 vehicle hijackings a day. This is a 30\% increase in comparison with the pre-Covid time of 2019 which reported a total of 17,777 hijackings \cite{b1}. It is, therefore, very clear that there is a staggering growth of vehicle hijackings in South Africa, which is a reasonable cause of concern.

In the technological and online world of today, social media platforms are the centerpiece of information exchange between people. One of largest and more notable platforms, Twitter, is an ever expanding means of communication and data generation due to the constant flow of user-to-user information. Users exchange strings of information on topics such as their own personal views, comical anecdotes, and most notably, current events. It is also used by news outlets to publish notable events in a timely manner. This, therefore, leads to the possibility of using this information to observe criminal events to assist in ensuring the safety of the general public, more specifically to the vehicle hijacking problem. Work done, such as that from Wang et al. \cite{b2} proposed using Twitter as a source real-time traffic event detection, whereby they were successful in their implementation. This provides further backing to the usage of Twitter to identify hijacking incidents.

The major issue with the incorporation of Twitter, is the identification of relevant and irrelevant tweets. In \cite{b3}, supervised learning techniques were proposed for this task; Convolutional Neural Network(CNN) achieved a favourable accuracy of 99.66\%. However, the use of supervised learning may not be feasible in real-time environments as issues of generalization of the classification may arise because the system will continuously encounter new types of data. Therefore, this work is aimed at detecting irrelevant tweets by unsupervised learning techniques for anomaly detection.

Goldstein and Uchida \cite{b4} performed an evaluation of 19 different unsupervised anomaly detection algorithms on 10 different datasets from varying application domains. Their aim was to provide a well-funded basis for unsupervised anomaly detection. For their evaluation, they incorporated an Area Under Curve (AUC). From their findings, when observing nearest-neighbour methods, the KNN method was found to be a top performer, mainly when given global anomaly detection tasks. It was also noted that from the cluster-based methods, the CBLOF provided poor performance. This was assumed to be due to the usage of weighting scores based on the number of members in each cluster, especially when given a small dataset. They resolved this by removing the weighting variable in a proposed uCBLOF method, which provided favourable results \cite{b4}. These techniques were recently employed anomaly detection in water meter readings in \cite{b18}.

Bibi et al. \cite{b5} proposed a novel unsupervised learning framework which is based on hierarchical and concept-based clustering for Twitter sentiment analysis. The hierarchical clustering methods single linkage, complete linkage, and average linkage algorithms are incorporated. For the feature representation methods, the 2 investigated were the Boolean, and Term Frequency-Inverse document frequency (TF-IDF). To ensure that the comparison is done fairly, the well-known classifiers, Naïve Bayes, and Neural Network are employed. To observe the performance, the accuracy performance metric is used. Their findings indicated that the unsupervised methods should be further explored due to the removal of required labelled data. For the feature representation, the TF-IDF was also found to have superior performance to the Boolean method \cite{b5}. The findings of their research, gives further backing into the exploration of unsupervised learning, as well as giving a further look into the feature extraction methods.

Devi and Saharia \cite{b6} proposed a mechanism for categorizing tweets based on their content by using the statistical and semantic features. They implemented the TF-IDF similar to the work by Bibi et al. \cite{b5}; it, however, was employed alongside a synonym based weighting scheme. The vector output of the TF-IDF was fed as an input for the synonym-based weighting scheme to generate a new vector. Both of these vectors were then used as a feature for clustering \cite{b6}. From their work, they were able to form 6 clusters, each addressing different categories, as well as obtaining a Silhouette coefficient score of 0.47 \cite{b6}. This work creates further backing into the usage of TF-IDF for the feature extraction used for unsupervised clustering.

Work such as that in \cite{b5} and \cite{b6} gave the inspiration for the work presented in this paper. As mentioned, the work that used supervised learning is not ideal for real-time implementation. Therefore, the work in \cite{b5} gave further insight into the usage of unsupervised learning on Tweet data. With the adoption of unsupervised learning, \cite{b4} presented the possibility of incorporating anomaly detection. This proved interesting as the relevant tweets could be treated as an individual cluster based on data similarities, and the irrelevant tweets would be outliers. To obtain the features for the clustering, the usage of the TF-IDF was also presented \cite{b5, b6}. Therefore the work in this paper would be exploring the usage of the nearest neighbour anomaly detection algorithm KNN, as well as the clustering based CBLOF algorithm. However, as mentioned due to the usage of a single large cluster, this work will be using an unweighted approach similarly to that presented in \cite{b4}. For the feature extraction process on the tweet data, this work will be using the TF-IDF. The data used in the formation of our cluster would be the relevant tweets from our 296 tweet dataset (76 tweets).

The work in this paper is organised as follows. A brief overview of the Anomaly Detection algorithms is presented in Section ~\ref{Section:Breif overview}. This is then followed by Section ~\ref{Section: Materials and Methods} which addresses the Materials and Methods employed in the implementation. The results and a discussion will be presented in Section ~\ref{Section: Results and Discussion}, before the conclusion in Section ~\ref{Section: Conclusion}.

\section{Brief Overview of the Anomaly Detection Algorithms Employed in this Work} \label{Section:Breif overview}
This section presents a brief overview of the two anomaly detection algorithms used in this work.

\subsection{K-Nearest Neighbour}
In the K-Nearest Neighbour algorithm, for each new point, the \textit{k} nearest data points are identified \cite{b7}. However for global anomaly detection there are other definitions. The current work focuses on the definition whereby the average distance to all of the k-nearest neighbours are calculated \cite{b8}. For this unsupervised method, the data points would be used to form clusters, whereby each point \textit{p} would be taken and then the sum of its distances to the k-nearest neighbours would be calculated to obtain a weight value, which is denoted as the weight of $p$, $w_k(p)$ \cite{b8}. Outliers would then be those points which have the largest weight values. Similar to the definition, the implementation in this work would be to calculate the sum of distances of each point $p$ to every other point $q$, and then obtain the mean average of all of the distances. The distance between two points is calculated through the Euclidean distance formula \cite{b9} as follows:
\begin{equation} \label{eq: Euclidean Distance}
D(p,q)= \sum_{i=1}^{N} (p_i-q_i)^2\,
\end{equation}

For the identification of an outlier, a threshold is implemented \cite{b4,b10}. If a point obtains an average distance greater than the threshold, it is then seen as an outlier. 

\subsection{Cluster Based Outlier Factor}
The Cluster Based Outlier Factor (CBLOF) is a method whereby each point is given an outlier factor, which is defined by the size of the cluster the point belongs to, as well as the distance to the closest cluster, granted the point currently belongs to a small cluster \cite{b11}. For the determination of a the cluster sizes (large or small), a large cluster would generally have a large majority of the data, corresponding to a selected threshold value, whereas a small clusters would hold the the minority of data points. By definition, suppose $C = {C_1, C_2, ... , C_k}$ is a set of clusters, such that $|C_1|\geq|C_2|\geq...\geq|C_k|$ \cite{b11}. Then given two parameters $\alpha$ and $\beta$, the following conditions are formed:
\begin{equation} \label{eq: ClusterForm1}
(|C_1|+|C_2|+...+|C_b|)\geq |D|*\alpha
\end{equation}
\begin{equation} \label{eq: ClusterForm2}
|C_b|/|C_(b+1)|\geq \beta
\end{equation}

If one of the above formulae holds, then the sets of the large and small clusters are defined as
\begin{equation}
    LC = {C_i|\leq b}
\end{equation}
\begin{equation}
    SC={C_j|j>b},
\end{equation}
where $b$ is the boundary \cite{b11}. In \eqref{eq: ClusterForm1}, if $\alpha$ is set to 70\%, this would mean that the large clusters would contain 70\% of the data. For \eqref{eq: ClusterForm2}, it is assumed that there would be a size difference between clusters corresponding to the $\beta$ value. For example, if $\beta$ was set to 4, the size of the large cluster would be at least 4 times greater than the size of the small cluster \cite{b11}.

For the determination of the outliers, data points are given an \textit{outlier score}, which is based on 2 basic rules \cite{b12}. Firstly, if a datapoint is a member of a large cluster, the calculated distance would be the distance to the centroid of its own cluster, and the corresponding outlier score would be that distance multiplied by the total number of data points in the cluster. Secondly, if a data point is a member of a small cluster, the calculated distance would be the distance to the centroid of the nearest large cluster, and the corresponding outlier score would be that distance multiplied by the total number of data points in its current small cluster \cite{b12}.

The implementation in this paper would be treating the entire relevant tweets dataset as a large cluster. Any point with a distance to the centroid of the cluster greater than a calculated threshold, would be deemed as an outlier.

\section{Materials and Methods} \label{Section: Materials and Methods}

\begin{figure}[htbp]
    \centerline{\includegraphics[scale=0.6]{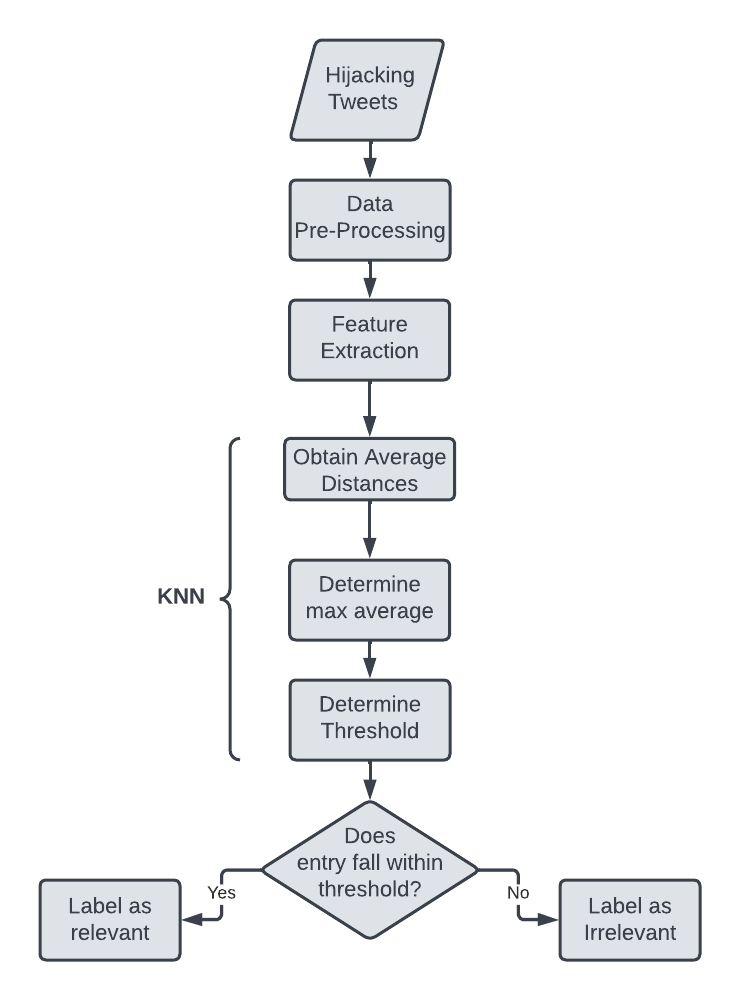}}
    \caption{KNN System Flow Chart}
    \label{Flowchart}
\end{figure}

\begin{figure}[htbp]
    \centerline{\includegraphics[scale=0.6]{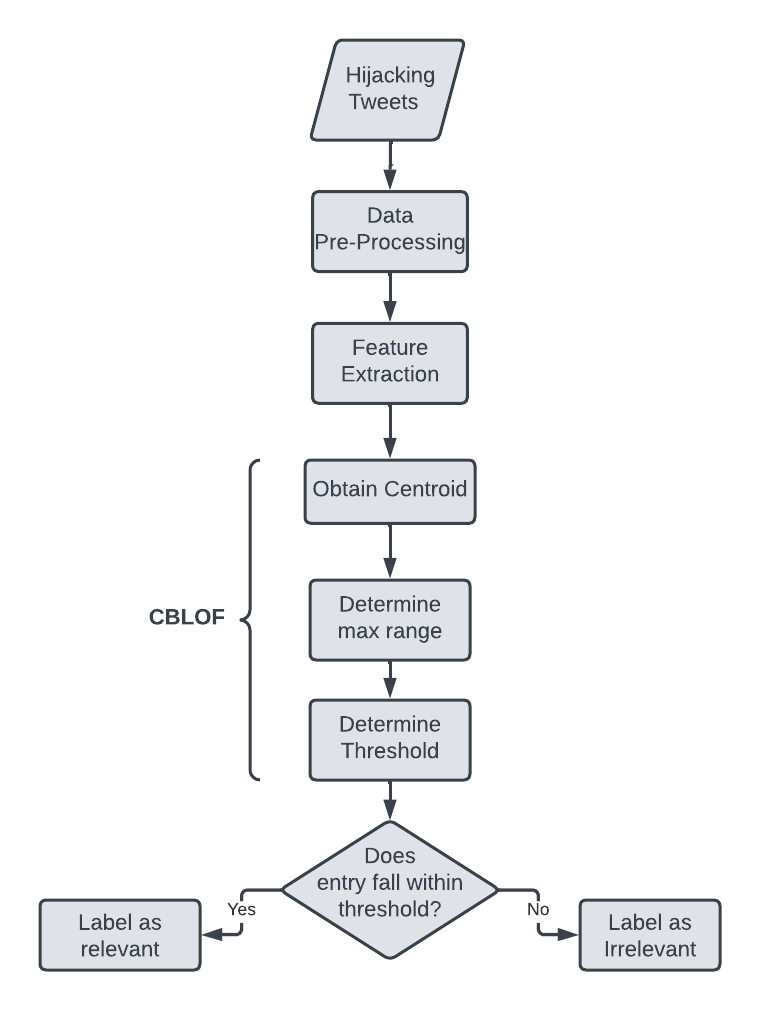}}
    \caption{CBLOF System Flow Chart}
    \label{CBLOF Flowchart}
\end{figure}

This section focuses on the materials and methods used in this paper to achieve the required output. The first step of the presented work, would be the cleaning and pre-processing of the Tweet data. Afterwards, feature extraction is applied to obtain the matrices used for the anomaly detection. The test data is then pushed through the 2 different anomaly detection methods separately with varying threshold values to observe performance. The following subsections will be detailing each of these processors. System flow diagrams for each method are presented in Figs.~\ref{Flowchart}-~\ref{CBLOF Flowchart}.

\subsection{Data Generation and Pre-processing}

The main focus of this work, is the usage of raw tweet data for the determination of the relevant and irrelevant tweets with regards to hijacking reports. Tweets are informal in nature, allowing uses to discuss anything they would like. This therefore brings forth the requirement to screen these tweets to identify the relevant, topical tweets required. For the work in this paper, the City of Cape Town was selected as the target area. The Tweets obtained would contain the keyword "hijacking" within a 50km radius of the center of Cape Town. These tweets were previously obtained using the Tweepy \cite{b13} python library. The formulated dataset contained 426 tweet entries. These tweets were then manually labeled to identify the relevant and irrelevant tweets using codes 1 and 0 respectively. They were then split into 296 tweets for the training set (76 relevant and 220 irrelevant), and 130 tweets for the testing set (29 relevant, 101 irrelevant). As previously stated, for the work done in this paper, the relevant set is used as our singular cluster due to this data being seen as the standard of relevant tweets. 

Before performing the feature extraction, data pre-processing is required. The Tweets are fed through a few pre-processing steps, such as reducing to lower case, removal of stop words and trailing spaces, etc. However, in contrast to the previous work done with the supervised learning methods \cite{b3}, removal of words such as "hijacking" are also done. This is due to these words shifting the weighting values during the feature extraction as they would naturally be present in all entries, relevant or irrelevant. Once the data has been pre-processed, as inspired by the work in \cite{b6} and \cite{b5}, the TF-IDF Vectorizer is employed. The TF-IDF Vectorizer grants a weighting value to each word in the document, based on the number of occurrences of a word in a document, as well as the number of occurrences of the word in the whole set of documents \cite{b14}. This is defined though the following equations:

\begin{equation} \label{eq: TF EQ}
TF = \frac{\text{number of occurrences of the term in the document}}{\text{number of terms in the document}}
\end{equation}

\begin{equation} \label{eq: IDF EQ}
IDF = \log\frac{\text{total number of documents}}{\text{number of documents containing the term}}
\end{equation}

\begin{equation} \label{eq: TFIDF EQ}
TF-IDF = TF * IDF
\end{equation}

\subsection{Implementation}
The work done in this paper was conducted in python 3.8, on a windows 10 system with an RTX 3060 Ti GPU. Once the relevant tweets were processed through the TF-IDF Vectorizer, any newly introduced data would be fed through each method separately to obtain the performance. The performance metrics used in this paper were influenced by the work done in \cite{b15, b16}. They are presented in the following equations:
    
\begin{equation} \label{eq: Accuracy}
Accuracy = \frac{\text{No. of correct predictions}}{\text{Total size of the dataset}}
\end{equation}

\begin{equation} \label{eq: Precision}
Precision = \frac{\text{True Positives}}{\text{True Positives + False Positives}}
\end{equation}

\begin{equation} \label{eq: Recall}
Recall = \frac{\text{True Positives}}{\text{True Positives + False Negatives}}
\end{equation}

\begin{equation} \label{eq: F1-Score}
F1-Score = \frac{\text{Precision x Recall}}{\text{Precision + Recall}}
\end{equation}

The K-Nearest Neighbour and the Cluster Based Outlier Factor methods are implemented as follows:
\begin{enumerate}
    \item \textit{K-Nearest Neighbour}: As mentioned in \cite{b4}, the KNN method is focused on taking a point, and observing the status of its \textit{k} nearest neighbours and basing a classification off of the found status. For anomaly detection however, a further definition discussed observing the average distances between \textit{k}-nearest neighbours \cite{b8}. For the method presented in this work, all of the relevant tweets would be identified as a single cluster. Therefore the approach of observing a specific $k$ value of points would not be ideal due to the possible disregard of the rest of the points in the cluster. Due to this, the $k$ value would be the total number of points in the cluster, and each new point would be compared to every other point in the cluster through the use of \eqref{eq: Euclidean Distance}. The obtained distances would then be used to identify the overall average distance between points in the cluster. To determine if a value is an outlier, the average distance of the new point to every point in the cluster would need to fall below an acceptable threshold. The threshold is calculated as 

    \begin{equation}\label{eq: knnthreshold}
    threshold = (1+\alpha).max(average)
    \end{equation}
    
    where $\alpha$ is a determined alpha value, and the $average$ is the list of average distances of each point in the relevant cluster to every other point in the cluster.
    
    \item \textit{Cluster Based Local Outlier Factor}:As described in \cite{b11, b12}, for the CBLOF implementation there is a large focus on the formation of the large and small clusters, and then the identification of the centroids. As mentioned earlier in the paper, for the current work, the entire relevant set would be used to form the large cluster. The \textit{outlier score} would therefore be based on the distance to the centroid of the relevant data cluster. The distance would be obtained using \eqref{eq: Euclidean Distance}. A point will hence be seen as an outlier, if the identified distance to the centroid, would be greater than an acceptable threshold value. The threshold would similarly to \eqref{eq: knnthreshold}, be calculated as 

    \begin{equation}\label{eq: threshold}
    threshold = (1+\beta).max(distance)
    \end{equation}

    where $\beta$ would be a determined beta value, and the distance would be the distance of all the data points in the relevant cluster to the centroid.
    
\end{enumerate}

For both, the KNN and the CBLOF implementations, the $\alpha$ and $\beta$ values in their respective threshold formulas (\eqref{eq: knnthreshold}, \eqref{eq: threshold}) will need to be selected. The process and selection will be presented in the Results and Discussion below.

\section{Results and Discussion} \label{Section: Results and Discussion}
In this section, the process of determining the correct threshold variables will be presented, along with the corresponding results. Following, will be the results and comparison between the KNN and CBLOF methods. The results will be presented using the previously mentioned performance metrics (Accuracy, Precision, Recall, and F1-Score).

\subsection{Setup and Determination of Alpha/Beta}
\begin{figure}[htbp]
    \centerline{\includegraphics[scale=0.7]{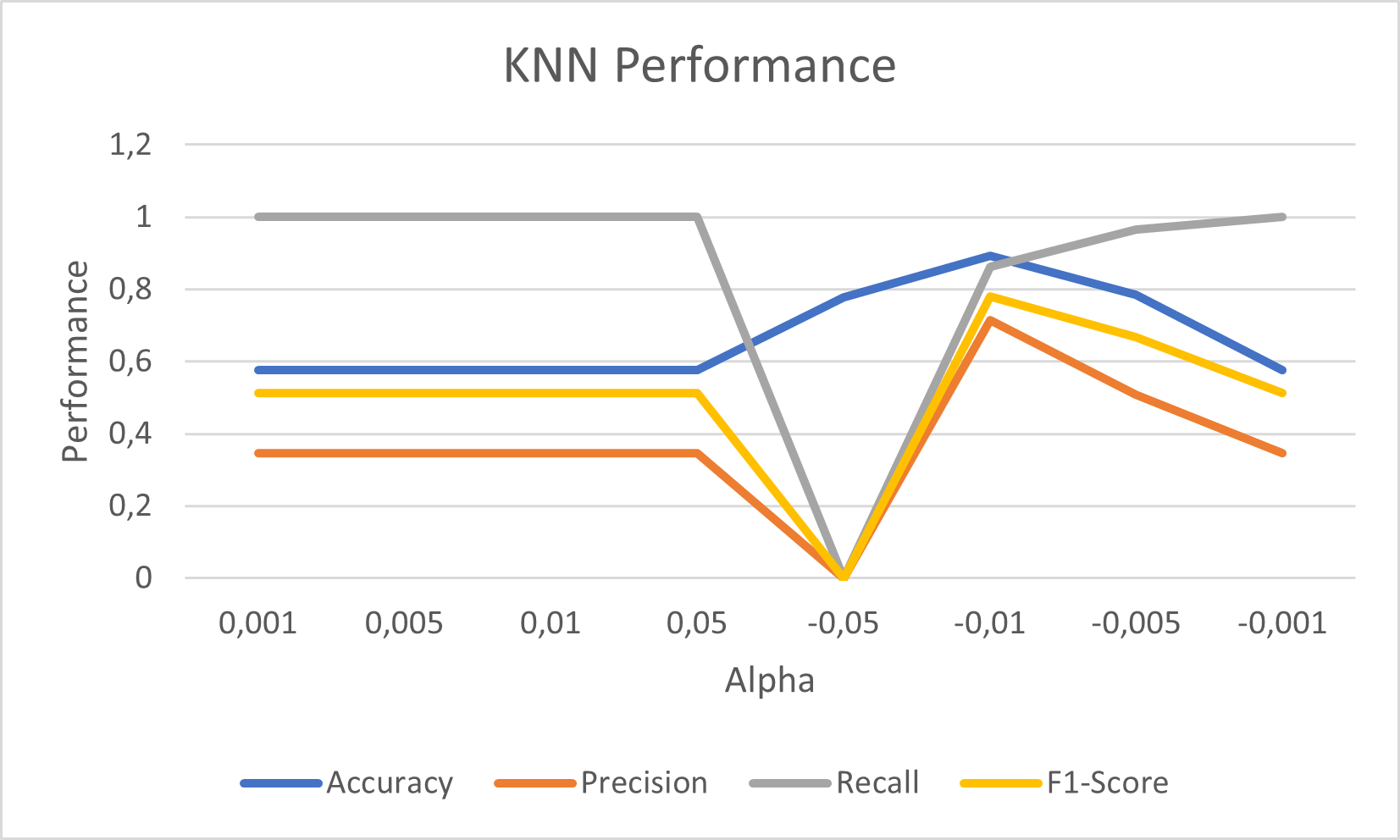}}
    \caption{KNN Performance Line Graph.}
    \label{KNNFlowchart}
\end{figure}

\begin{figure}[htbp]
    \centerline{\includegraphics[scale=0.7]{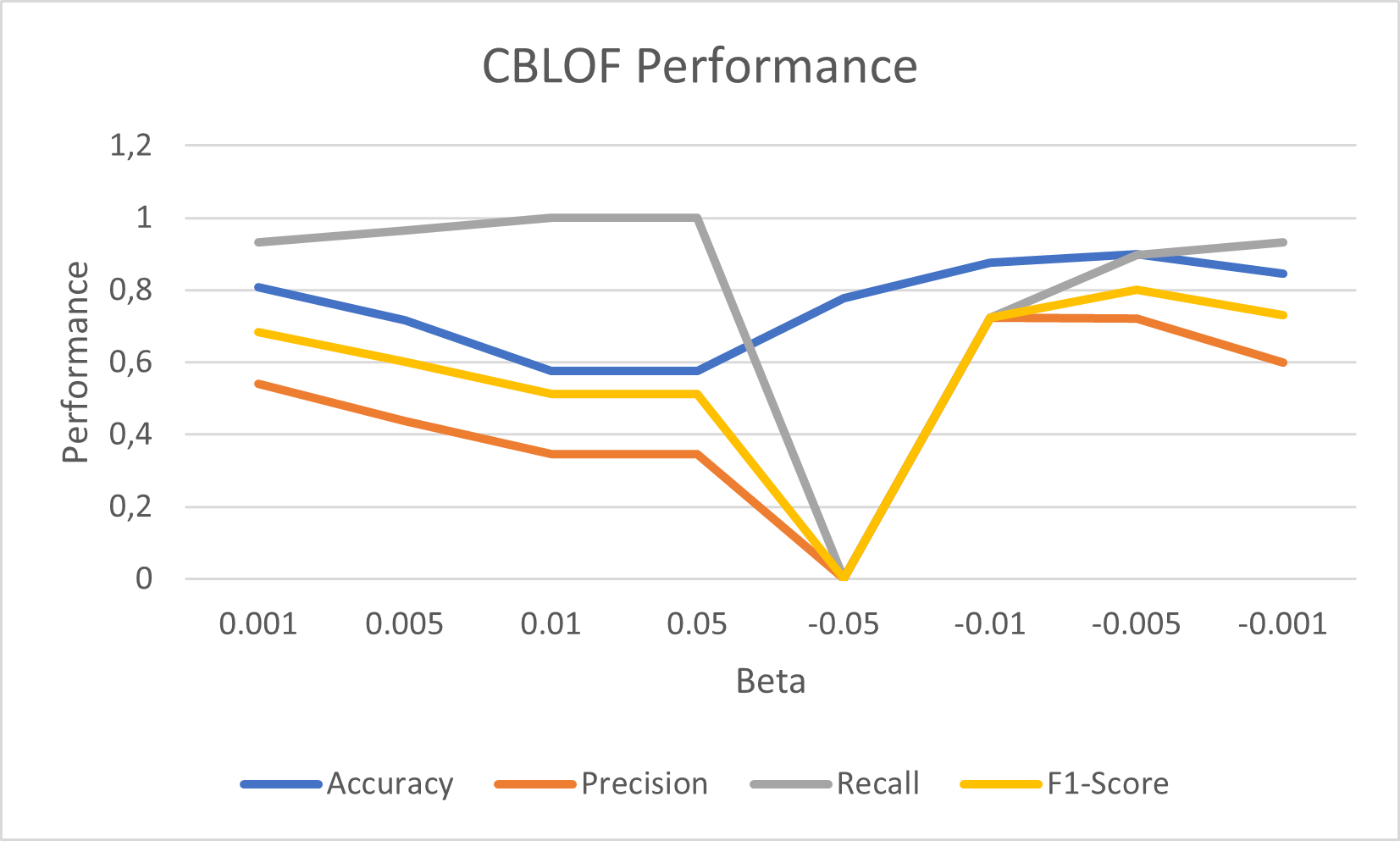}}
    \caption{CBLOF Performance Line Graph.}
    \label{CBLOFFlowchart}
\end{figure}

As discussed, the $\alpha$ and $\beta$ values used in the threshold calculations would need to be selected. The determination of these values are identified through sufficient testing. Firstly, for these values, a range would need to be identified. Therefore, a selection of 8 possible values were chosen: 0.001, 0.005, 0.01, 0.05, -0.001, -0.005, -0.01, and -0.05. The usage positive and negative values were employed to observe the ways in which the increase and decrease of the threshold would affect the overall performance of the algorithms. After the implementation of each method, a series of test data is fed into each method. Through the use of each $\alpha$/$\beta$ value, a threshold value is calculated and the classification of the relevant data and irrelevant data(outliers) is done. Due the usage of labeled data to observe performance, the presented method is therefore a semi-supervised approach. The performance of the varying threshold values were stored, and they are presented in Table ~\ref{KNNTab} and ~\ref{CBLOFTab}, along with the graphical representations shown in Figs.~\ref{KNNFlowchart}-~\ref{CBLOFFlowchart}.

\begin{table}[htbp]
\caption{KNN Performance Comparison}
\begin{center}
\begin{tabular}{ |p{1cm}|p{1cm}|p{1cm}|p{1cm}|p{1cm}|p{1cm}| }
 \hline
 \multicolumn{6}{|c|}{KNN Performance with varying Alpha values} \\
 \hline
  $\alpha$& Threshold & Accuracy &Precision &Recall &F1-Score\\
 \hline
  0.001 & 1.415628   &0.576923& 0.345238& \textbf{1.000000}& 0.513274\\
  0.005 & 1.421285 & 0.576923& 0.345238&  \textbf{1.000000}& 0.513274\\
  0.010 & 1.428356& 0.576923&  0.345238& \textbf{1.000000}& 0.513274\\
  0.050 &1.484924 & 0.576923&  0.345238&  \textbf{1.000000}& 0.513274\\
  -0.050 &1.343503 & 0.776923&  0.000000& 0.000000& 0.000000\\
  -0.010 &1.400071 & \textbf{0.892308}&  \textbf{0.714286}& 0.862069& \textbf{0.78125}\\
  -0.005 &1.407142 & 0.784615& 0.509091&  0.965517& 0.666667\\
  -0.001 &1.412799 & 0.576923& 0.345238& \textbf{1.000000}& 0.513274\\
 \hline
\end{tabular}
\label{KNNTab}
\end{center}
\end{table}

\begin{table}[htbp]
\caption{CBLOF Performance Comparison}
\begin{center}
\begin{tabular}{ |p{1cm}|p{1cm}|p{1cm}|p{1cm}|p{1cm}|p{1cm}| }
 \hline
 \multicolumn{6}{|c|}{CBLOF Performance with varying Beta values} \\
 \hline
  $\beta$& Threshold & Accuracy &Precision &Recall &F1-Score\\
 \hline
  0.001 & 1.004428   &0.807692& 0.540000& 0.931034&0.683544\\
  0.005  & 1.008441 & 0.715385& 0.437500& 0.965517& 0.602151\\
  0.010 & 1.013458& 0.576923&  0.345238& \textbf{1.000000}& 0.513274\\
  0.050 &1.053595 & 0.576923&  0.345238&  \textbf{1.000000}&0.513274\\
  -0.050 &0.953253 & 0.776923&  0.000000& 0.000000& 0.000000\\
  -0.010 &0.99339 & 0.876923& \textbf{0.724138}&0.724138& 0.724138\\
  -0.005 &0.998407 & \textbf{0.900000}& 0.722222&  0.896552&\textbf{0.800000}\\
  -0.001 &1.002421 & 0.846154&0.60000&  0.931034&0.729730\\
 \hline
\end{tabular}
\label{CBLOFTab}
\end{center}
\end{table}

As seen in both figures, when using the $\alpha$ or $\beta$ values of -0.05, the performance takes a large dip. When observing the resultant data to identify what occurred during this process, it seemed that all the predictions were marked as anomalies (Prediction = 0). This is also indicated by the accuracy of 0.776923 which corresponds to the ratio of irrelevant tweets in our test set (101 out of 130). Therefore, the usage of -0.05 should be disregarded for both methods. 

Now looking at Table ~\ref{KNNTab}, an interesting outcome in the 4 positive $\alpha$ values, as well as with $\alpha$ -0.001 was identified, whereby they all returned the exact same outcomes along with a Recall of 1. This indicates that these $\alpha$ values provided 0 false negatives. It is assumed that this is due to the large threshold, resulting in the all actual relevant tweets being marked as relevant, however as well as allowing a large number of irrelevant tweets (anomalies) into the cluster as well. This assumption is reinforced by the $\alpha$ values -0.01 and -0.005 which had a considerable leap in Accuracy, as well as overall performance. This is hence assumed to be due to the reduction of the threshold, avoiding an excess number of irrelevant tweets being accepted as relevant entries. Therefore, based on the results presented, -0.01 is selected as the preferred $\alpha$ value.

When observing Table ~\ref{CBLOFTab}, a similar situation as presented in the KNN results is seen. When using  $\beta$ values of 0.01 and 0.05, the identical results to those presented previously are found. This can be once again assumed to be due to the larger threshold value and should therefore be disregarded. The highest accuracy identified is when using $\beta$ -0.005, which obtained an accuracy of 0.9\%. Similar to the KNN method, the $\beta$ of -0.01 performed well with an accuracy of 0.876923, as well obtaining the highest precision value of 0.724138. When using $\beta$ -0.005, there was only a slight decrease in the Precision in comparison to -0.01, however it achieved a notably higher recall, which resulted in a considerable increase in the F1-Score. This can also be visually seen in Fig. ~\ref{CBLOFFlowchart}, where there is an observable increase in performance with $\beta$ -0.005. Therefore, there is enough backing to select -0.005 as the preferred $\beta$ value for the CBLOF method.

\subsection{Comparison Between Methods}
\begin{figure}[htbp]
    \centerline{\includegraphics[scale=0.7]{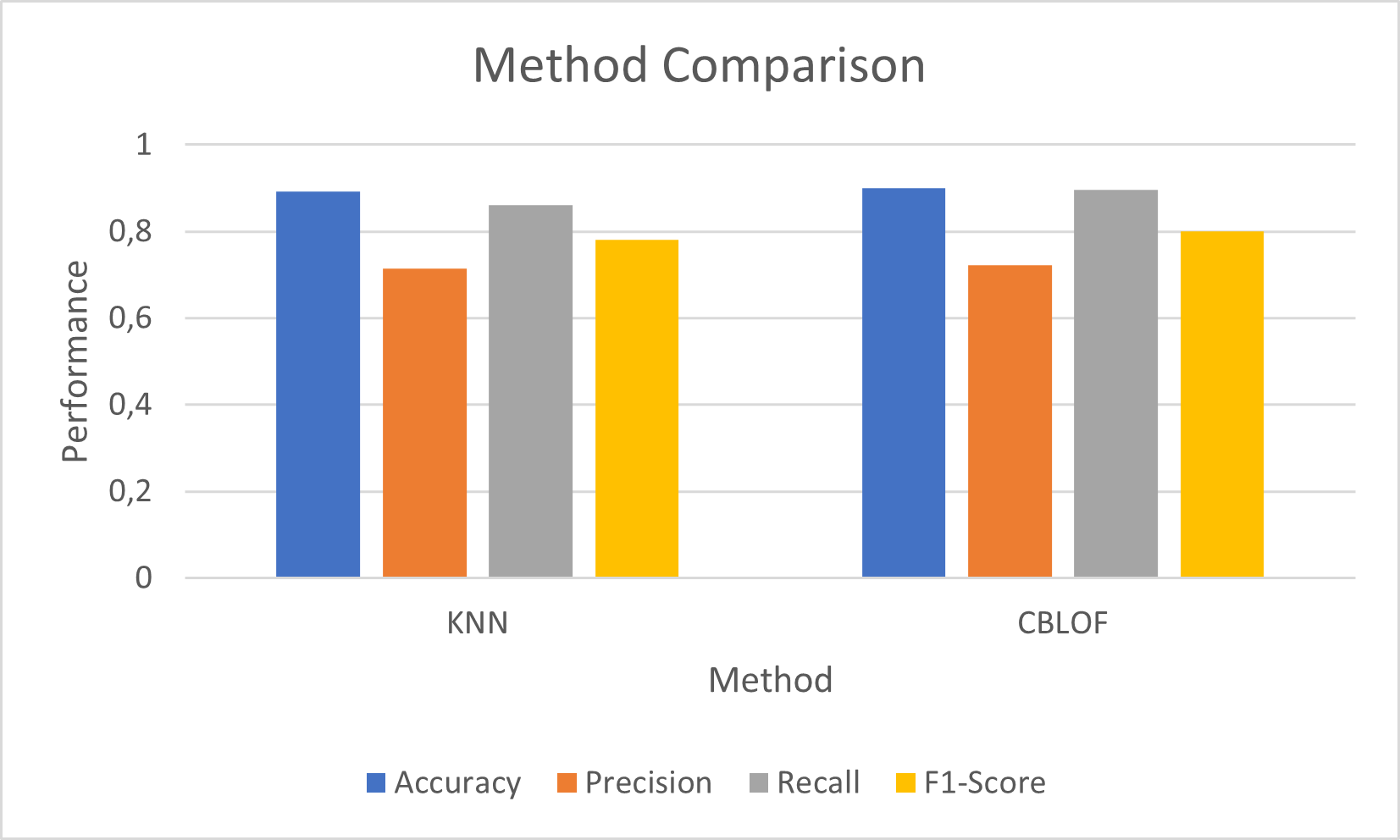}}
    \caption{Method Comparison Bar Graph.}
    \label{MethodComparison}
\end{figure}

After obtaining the results of the varying $\alpha$ and $\beta$ values, and observing the best performing thresholds (-0.01 $\alpha$ and -0.005 $\beta$), a comparison between the 2 methods should be done. The result of this comparison is presented in Table ~\ref{KNNCBLOFTab}, as well as a visual representation in Fig. ~\ref{MethodComparison}. 
\begin{table}[htbp]
\caption{Method Performance Comparison}
\begin{center}
\begin{tabular}{ |p{1cm}|p{1cm}|p{1cm}|p{1cm}|p{1cm}|p{1cm}|p{1cm}| }
 \hline
 \multicolumn{6}{|c|}{Comparison of KNN and CBLOF} \\
 \hline
Method& $\alpha$/$\beta$ & Accuracy &Precision &Recall &F1-Score\\
 \hline
  KNN& -0.01& 0.892308& 0.714286& 0.862069&0.78125\\
  CBLOF& -0.005& \textbf{0.9}& \textbf{0.722222}&  \textbf{0.896552}&\textbf{0.8}\\
 \hline
\end{tabular}
\label{KNNCBLOFTab}
\end{center}
\end{table}

As presented, both methods performed considerably well. The KNN method, as well as the CBLOF method obtained high Recall values. This indicates that both methods were able to excel when identifying the number of actual relevant tweets while reducing the number of false negatives. For the system presented in the current work of identifying hijacking reports, this is a highly desirable metric. When observing the rest of the performance metrics, the CBLOF method was able to achieve a slightly greater performance in all aspects, however minuscule it may seem. The overarching results are comparable with the work done in \cite{b17}, whereby they found the nearest-neighbour algorithm to perform slightly better in 2 out of 3 datasets, however the clustering based algorithms, including the unweighted CBLOF had a superior performance in the 3rd dataset. Therefore, it can be said that both the KNN and CBLOF methods would be highly applicable, however the complexity of the methods should be taken into account as the clustering method would generally have a lower complexity \cite{b17}. Hence, for the current work, the CBLOF method is selected as the preferred algorithm.

\section{Conclusion} \label{Section: Conclusion}

The work presented was focused on employing an semi-supervised approach for the determination of hijacking tweets. It therefore undertook the task of incorporating anomaly detection based algorithms due to the nature of the problem, whereby an irrelevant tweet would be seen as an anomaly/outlier. Hence, two anomaly detection methods were selected: K-Nearest Neighbour, and the Cluster Based Local Outlier Factor. For this approach, 76 relevant tweets were extracted from a dataset of 296 tweets, to formulate a single relevant cluster. Using the produced cluster, both models were employed and the outlier detection methods were optimised by introducing a threshold value. Through the determination of the best performing thresholds for each method, an $\alpha$ of -0.01, and $\beta$ of -0.005 were selected respectively. When performing a final comparison between both methods, it was found that the CBLOF method was able to achieve a slightly greater performance in all areas. This result, along with the consideration of complexity, lead to the selection of the CBLOF as the preferred method. The findings in this work presents a strong argument for the usage of unsupervised anomaly detection when identifying relevant hijacking tweets.

For future work, the comparison between the anomaly detection methods and the supervised learning methods would be addressed. As well as incorporating further optimisation with the aim of increasing the performance.

\bibliographystyle{IEEEtran}
\bibliography{SATNAC_Latex_Template_2017}

\vspace{-1.0cm}

\begin{IEEEbiographynophoto}{Taahir Aiyoob Patel}
received his Bachelors Degree (Honors) in Computer Science from the University of the Western Cape. He is currently working on his MSc in Computer Science at the University of the Western Cape. His research interest is in Machine Learning, with a more direct focus on Text Classification and its use on data obtained from Social Media platforms. 
\end{IEEEbiographynophoto}

\vspace{-1.0cm}

\begin{IEEEbiographynophoto}{Clement N. Nyirenda}
received his PhD in Computational Intelligence from Tokyo Institute of Technology in 2011. His research interests are in Computational Intelligence paradigms such as Fuzzy Logic, Swarm Intelligence, and Artificial Neural Networks and their applications in Communications.
\end{IEEEbiographynophoto}
\vfill

\end{document}